\documentclass[12pt, oneside]{article}
\usepackage[letterpaper, left=1in, right=1in, top=1in, bottom=1in]{geometry}
\usepackage{setspace} 
\usepackage{lineno} 
\usepackage{amssymb}
\usepackage{amsmath}
\usepackage{dsfont}
\usepackage{mathptmx} 
\usepackage{graphicx}
\usepackage[labelfont=bf]{caption} 
\usepackage{float} 
\usepackage{booktabs}
\usepackage{multirow}
\usepackage[ 
	style=authoryear,
	citestyle=authoryear,
	maxbibnames=6,
	giveninits=true,
    uniquename=init,
	backend=bibtex,
	url=false,
	doi=false,
	isbn=false,
	eprint=false
	]{biblatex}
\AtEveryBibitem{%
\clearfield{month}%
\clearfield{date}%
}
\usepackage{tikz}
\newcommand{\centersection}[1]{%
  \begin{center}
    \vspace{0.2cm}%
    \textbf{#1}%
    \vspace{-0.2cm}%
  \end{center}
}

\begin{document}


\pagenumbering{gobble}

\title{Histology-informed tiling of whole tissue sections improves the interpretability and predictability of cancer relapse and genetic alterations}
\author{
Willem Bonnaff\'e$^{1,2,3,*}$,
Yang Hu$^{2}$,
Andrea Chatrian$^{2}$,\\
Mengran Fan$^{2}$,
Stefano Malacrino$^{3}$,
Sandy Figiel$^{3}$,
CRUK ICGC Prostate Group$^{4}$,\\
Srinivasa R. Rao$^{3}$,
Richard Colling$^{3}$,
Richard J. Bryant$^{3}$,
Freddie C. Hamdy$^{3}$,\\
Dan J. Woodcock$^{3}$,
Ian G. Mills$^{3}$,
Clare Verrill$^{3}$,
Jens Rittscher$^{2}$
}
\date{}
\maketitle

\begin{center}
\vspace{-0.5cm}
$^{1}$Department of Biology, Life and Mind Building, University of Oxford, South Parks Road, Oxford OX1 3EL, UK\\
$^{2}$Institute of Biomedical Engineering, Big Data Institute, University of Oxford, Old Road Campus, Oxford OX3 7LF, UK\\
$^{3}$Nuffield Department of Surgical Sciences, University of Oxford, Old Road Campus Research Building, Oxford OX3 7DQ, UK\\
$^{4}$A list of members and affiliations appears in the acknowledgements.\\ 
$^{*}$Corresponding author
\end{center}

\centersection{Abstract}

Histopathologists establish cancer grade by assessing histological structures, such as glands in prostate cancer. Yet, digital pathology pipelines often rely on grid-based tiling that ignores tissue architecture. This introduces irrelevant information and limits interpretability. We introduce histology-informed tiling (HIT), which uses semantic segmentation to extract glands from whole slide images (WSIs) as biologically meaningful input patches for multiple-instance learning (MIL) and phenotyping. Trained on 137 samples from the ProMPT cohort, HIT achieved a gland-level Dice score of 0.83 ± 0.17. By extracting 380,000 glands from 760 WSIs across ICGC-C and TCGA-PRAD cohorts, HIT improved MIL models AUCs by 10\% for detecting copy number variation (CNVs) in genes related to epithelial-mesenchymal transitions (EMT) and \textit{MYC}, and revealed 15 gland clusters, several of which were associated with cancer relapse, oncogenic mutations, and high Gleason. Therefore, HIT improved the accuracy and interpretability of MIL predictions, while streamlining computations by focussing on biologically meaningful structures during feature extraction.

\vspace{0.5cm}
\noindent
\textbf{Keywords:}
Digital Pathology; Histology-Informed Tiling; Prostate Cancer; Multiple-Instance Learning; Cancer Phenotyping; Semantic Segmentation; Copy Number Variation; Biochemical Recurrence; Gleason Grading;

\vspace{0.5cm}
\noindent
\textbf{Email:}
w.bonnaffe@gmail.com

%


\newpage
\pagenumbering{arabic}
\setcounter{page}{1}
\setstretch{1.2}

\section{Introduction}

Cancer is the leading cause of death worldwide, with the most common being carcinomas of the breast, lung, colon, rectal, and prostate (\cite{Weinberg_2014}). Prostate cancer is the most frequently diagnosed malignancy in men, and a leading cause of cancer-related mortality worldwide. Its clinical management is complicated by the dual challenges of over-treatment of indolent disease, and under-treatment of aggressive disease, particularly in patients with intermediate-risk disease which has a variable clinical course. Over-treatment of indolent disease can result in unnecessary adverse effects without survival benefit, while under-treatment risks progression to an advanced, incurable metastatic stage (\cite{Bukowy_2020}). As the disease burden grows through a rising incidence of prostate cancer (\cite{James_2024}), accurate risk-stratification of patients with prostate cancer confined to the prostate gland based on disease aggressiveness has become a priority to ensure appropriate therapeutic interventions (\cite{Xie_2022}). 

Histopathologists establish the grade of cancer through scoring key histological features of tissue sections, or slides, under optical microscopy. In breast cancer, the Nottingham grading system involves scoring mitotic counts, nuclear pleomorphism, and tubular formation (\cite{Elston_1991, Jaroensri_2022}). In prostate cancer, the cornerstone of diagnosis and risk-stratification is the Gleason grading system, which assesses architectural patterns of cancer glands within magnified histological tissue sections (\cite{Gardner_1988}). This examination is difficult for intermediate-risk cancer, where differences in the morphological properties of the tissue between aggressive and benign cancer forms are less striking. In particular, intermediate-grade prostate cancer (i.e. Gleason 3+4 and 4+3) are often associated with considerable prognostic uncertainty (\cite{Zhang_2024}). Better cancer phenotyping tools that achieve nuanced characterisation of tissue morphologies, particularly glandular structures, could provide the basis for improving stratification and minimise clinical dilemmas associated with these borderline cases (\cite{Xie_2022}).

The digitisation of histopathology practices and the development of artificial intelligence hold great potential for analysing fine morphological features on unprecedented scales, thereby improving diagnosis and prognosis, and the detection of molecular biomarkers associated with aggressive cancers. Progress has been achieved in two distinct but complementary areas of machine learning.

The first area, semantic segmentation, enabled the automatic extraction of key histological features from whole slide images (WSIs), with a focus on nuclei and glandular structures (\cite{Nasir_2023}). These methods, often based on deep learning architectures like U-Net and its variants (\cite{Ronneberger_2015}), have shown promise in automating the segmentation task, paving the way for large-scale and consistent analyses of glandular structures (e.g. \cite{Ciga_2022, Zhang_2024}). This makes it possible to process entire cohorts of patients and gather histological structures for downstream tasks such as diagnosis and prognosis. Despite their utility, existing gland segmentation approaches typically stop at the gland identification stage and do not delve into the fine morphological characterisation of individual glands. As a result, these methods fall short in providing insights into subtle morphological variations that might underpin prognostic differences, particularly in intermediate-risk prostate cancer cases (\cite{Nasir_2023, Zhang_2024}).

The second area, multiple-instance learning (MIL), a weakly-supervised learning method, has emerged as the gold standard for analysing large WSIs (\cite{Isle_2018, Lu_2021}). Digitised whole tissue sections are initially divided following a grid into sets of patches referred to as bags. A pre-trained encoder then compresses each patch into a lower-dimensional embedding vector. Finally, an aggregator architecture collects information across all patches of a bag to make a prediction at the slide-level. Much work has been done on improving the different parts of this framework, for instance, by including tissue patches at multiple magnification scales (\cite{Shi_2023b}), fine-tuning the encoder using contrastive learning (CL) (\cite{Tavolara_2022, Liu_2024}), or improving the architecture of the aggregator by first including attention-based pooling (\cite{Isle_2018}) and clustering (\cite{Lu_2021}). Typical applications include diagnosis (\cite{Campanella_2019, delAmor_2022}), prognosis (\cite{Veldhuizen_2023}), or detecting molecular features (\cite{Bilal_2021, Tomita_2022, Erak_2023, Liu_2024, Deng_2023, ElNahhas_2024}) from tissue morphology.

Despite our increased capacity to segment histological structures and progress in these weakly-supervised learning approaches, very little work has focussed on integrating them. Most MIL approaches still rely on grid-tiling, whereby patches are extracted regardless of tissue structure. This results in the truncation of larger histological structures, such as glands, the shape of which is a key indicator of the disease grade and risk of disease progression in both breast (\cite{Jehanzaib_2025}) and prostate cancer (\cite{Serafin_2023}). Therefore, this system for tissue representation does not link well with the existing and traditional expert knowledge. This not only constrains the performance of the algorithms, but also reduces the interpretability of the outputs.

In this manuscript we address the critical question whether tiling the tissue according to histological structures, especially glands, by leveraging an instance segmentation approach, could improve the performance and interpretability of MIL models for three crucial tasks in digital histopathology: establishing a prognosis,  detecting genetic alterations, and cancer phenotyping. 

We consider biochemical recurrence (BCR) as a clinically relevant endpoint given that it is a common indicator of prostate cancer relapse and has been shown to correlate with Gleason grading (\cite{Yanagisawa_2008}). We hypothesise that there will be a strong association between gland morphology and BCR, and so a higher accuracy in predicting relapse when using histologically informed tiling around glands.  We also consider copy number variation (CNV) in genes controlling epithelial-mesenchymal transitions (EMT), such as \textit{ZEB1}, \textit{ZEB2}, \textit{SNAI1}, and \textit{SNAI2}, as they have been linked to Gleason grading and poor prognosis (\cite{Cheaito_2019}), and to gland morphology due to their destabilising effect on epithelial layers (\cite{Mak_2010}). Given this, we expect histologically informed tiling to improve the accuracy of the detection of CNV of EMT genes. Finally, we also investigate how attention scores at the level of histological structures, rather than arbitrary truncated tiles, can reveal the morphology of glandular structures associated with cancer relapse and EMT. We expect that in both cases, glands associated with poor prognosis and EMT should contain a higher number of cells and a less regular structure.

\section{Results}

To address these hypotheses, we propose a histology-informed tiling (HIT) approach for MIL and cluster-based phenotyping (Fig. 1). First, using a first cohort (ProMPT) we trained a segmentation model, \textit{GlandSeg}, to tile whole slide images around detected glands. We applied \textit{GlandSeg} to three datasets from two additional cohorts (ICGC-C and TCGA-PRAD) to generate three tile sets. The first tile set was paired with annotations from histopathologists for cluster-based phenotyping of cancer gland morphologies associated with different Gleason grades. The two remaining tile sets were used for the training of MIL models to predict cancer relapse and genetic alterations, assess performance gains over standard tiling approaches, and reveal the phenotype of associated cancer glands.  

\begin{figure}
\begin{center}
\includegraphics[width=\linewidth, page=1]{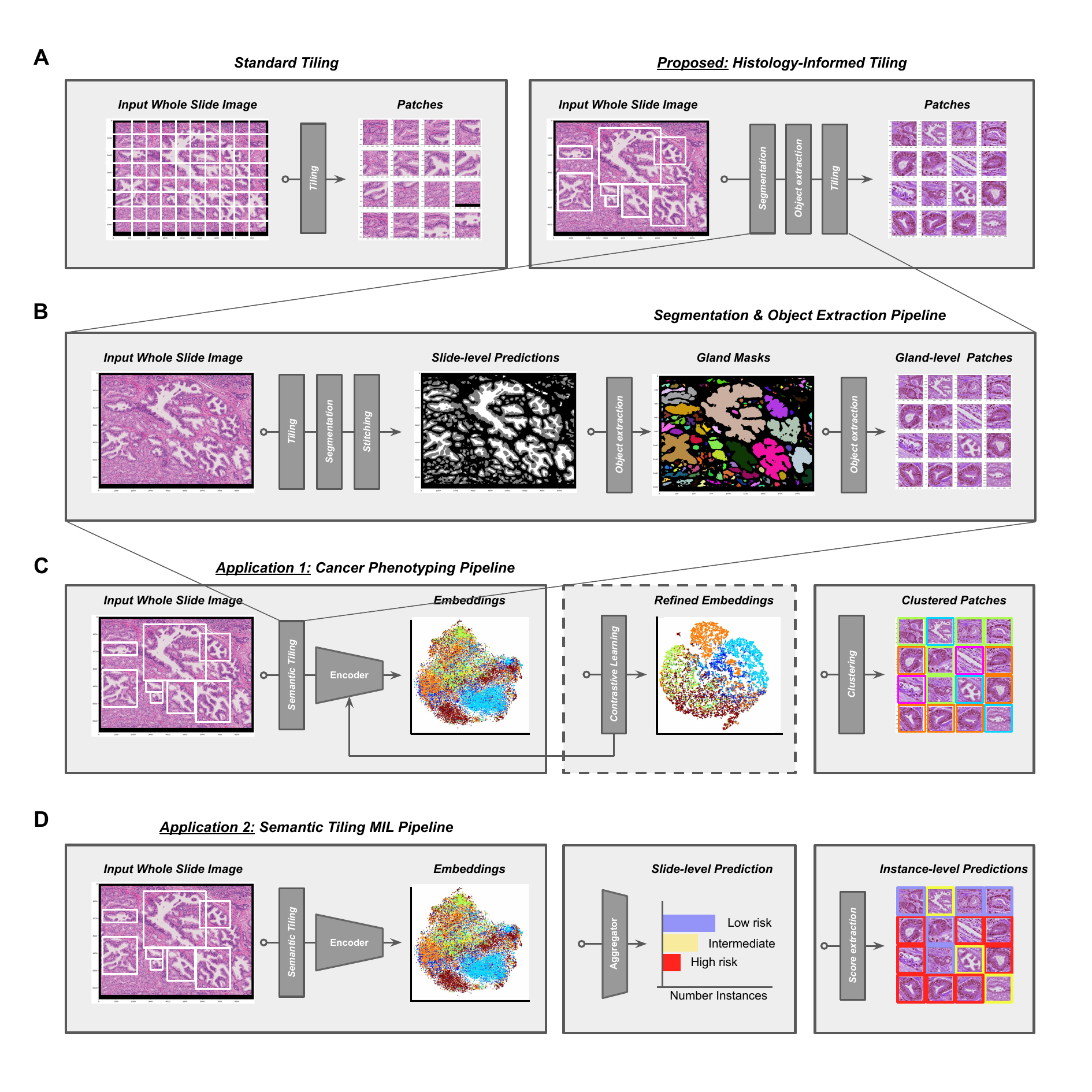}
\end{center}
\caption{
\textbf{Overview of the histology-informed tiling (HIT) framework and its applications.} Panel A shows the difference between grid tiling and HIT, also named semantic tiling. Panel B shows the steps required to perform semantic tiling, which involves semantic segmentation and object extraction. Panel C provides a first application for phenotyping cancer tissue morphology with dimensionality reduction, contrastive learning (where the dashed-line box indicates an optional step), and clustering. Panel D displays a second application in multiple instance learning based on patches extracted by HIT.
}
\end{figure}

\centersection{Accurate gland segmentation as the cornerstone of semantic tiling}

Our segmentation model, thereafter referred to by \textit{GlandSeg}, yielded high quality identification of glandular structures, as indicated by the high segmentation accuracy of stroma/background and of whole glands (Fig. 2A, stroma/background Dice score [mean ± sd] = 0.91 ± 0.12; gland Dice score = 0.83 ± 0.17), on par with state-of-the-art methods (e.g. \cite{Salvi_2021, Jehanzaib_2025}). The subdivision of whole glands into epithelium and lumen was also relatively accurate, albeit lower than that of detecting whole glands (Fig. 2A, epithelium Dice score = 0.68 ± 0.14; lumen Dice score = 0.71 ± 0.28). The fact that the epithelium compartment was detected with a relatively lower accuracy, compared to the other compartments, is not surprising, as the boundary between the stroma and epithelium is less clear than that between the epithelium and the lumen. Another study also found that the epithelium compartment was relatively harder to detect than the stromal compartment, but that lumen was the hardest of the three (\cite{Bukowy_2020}). 

While \textit{GlandSeg} was trained on tissue samples from 24 radical prostatectomy cases of the ProMPT cohort, we then applied it to three datasets derived from the two additional independent cohorts, ICGC-C and TCGA-PRAD (see methods for details). This resulted in the segmentation and extraction of 23,415 glands from 10 formalin-fixed paraffin-embedded (FFPE) WSIs of the Cancer Genome Atlas cohort for prostate adenocarcinoma (TCGA-PRAD), 107,199 instances from 516 WSIs of the Cambridge prostate adenocarcinoma cohort of the International Cancer Genome Consortium (ICGC-C), and 251,532 glands from the 234 WSIs of the TCGA-PRAD cohort embedded via optimal cutting temperature (OCT). Examples are shown in Fig. 2D.

Overall, this pre-processing step, which we refer to as semantic tiling, or HIT, sets a better foundation for downstream tasks than standard tiling by reducing the amount of information processed, while increasing its quality. It is more computationally efficient to encode glands, which account for a fraction of the tissue, than to blindly encode the whole tissue. It is also more biologically relevant to focus on glands, as this is how histopathologists score cancer grades. Other approaches have embedded location constraints at the feature extraction stage and demonstrated a gain in performance (\cite{delAmor_2022}). However, these methods still require the processing of the entire WSI, which is less efficient than encoding only the relevant histological structures.

\begin{figure}
\begin{center}
\includegraphics[width=\linewidth, page=2]{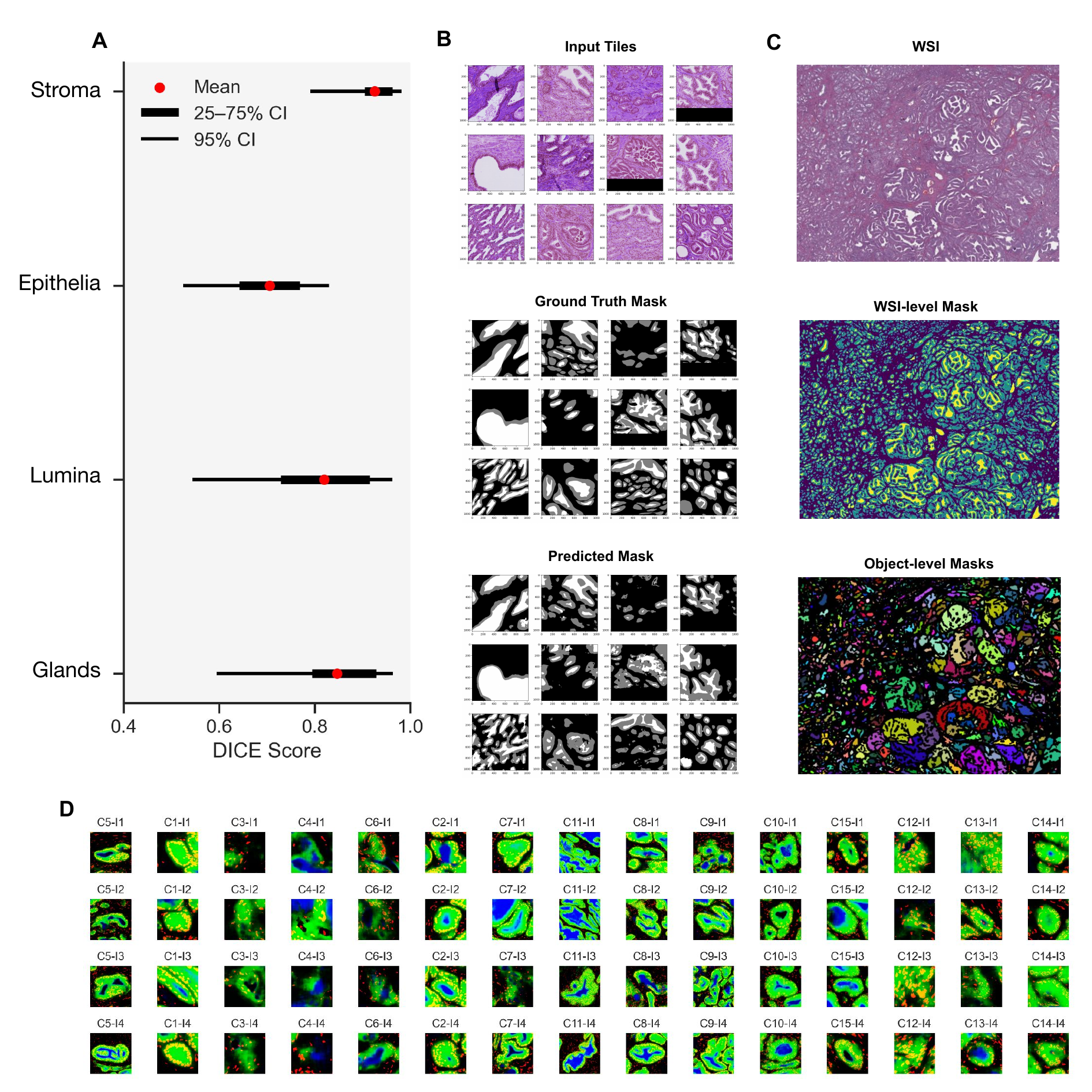}
\end{center}
\caption{
\textbf{Performance and examplar results from the segmentation and object extraction pipeline.} Panel A displays the accuracy (Dice score) of the model in segmenting different compartments of the tissue: stroma, epithelium, lumen, and glands (epithelium and lumen together). Panel B shows examples of predicted masks compared to ground truth annotations for randomly selected patches. Panel C demonstrates the slide-level mask obtained after stitching patch-level predictions and the object extraction step. Panel D shows examples of semantically-extracted tiles, where each patch contains an individual glandular structure. The background is coloured black, epithelia green, lumen blue, and nuclei red.
}
\end{figure}

\centersection{Gland clusters derived from HIT match cancer grading}

One of the immediate benefits of HIT is that each patch contains a distinct histological structure, i.e. glands in the case of prostate cancer. The corresponding embedding space therefore only contains histologically meaningful regions. We assessed how well the phenotyping, i.e. the morphological categorisation of glands, achieved by the machine aligns with the view of a trained histopathologist. 

To do this, we first compressed each extracted gland using convolutional neural networks (CNN) and vision transformers (ViT) encoder architectures and performed hierarchical agglomerative clustering (HAC) on the embeddings (N = 512) within each dataset. Overall, the intra-to-inter-cluster variance ratio indicated an optimal number of clusters above 15 for the three datasets (Fig. S1B). Fine-tuning the encoders through CL with triplet loss improved the segregation of the clusters by increasing the inter-cluster distance and reducing intra-cluster variance (Fig. S1C), resulting in clearer delineation of groups of histologically similar gland instances (Fig. S1B).

Fig. 3 provides a comparison of the morphological clusters identified through HAC with gland-level Gleason scores annotated by histopathologists. The annotations of histopathologists identified four distinct regions in the embedding space corresponding to high-grade (4+4 and above, Fig. 3B, in red), unfavourable intermediate (4+3, Fig. 3B, in orange), favourable intermediate (3+4, Fig. 3B, in yellow), and background (Fig. 3B, in gray). Clusters 12, 13 and 9 mapped to the high-grade region, while clusters 14, 15, and 10 mapped to the unfavourable intermediate region (Fig. 3A). The remaining clusters coincided well with the favourable intermediate and background regions. Interestingly, cluster 12 and 13, which were associated with higher Gleason patterns, also displayed a relatively higher nuclei density (Fig. 3D, orange and red distribution).

Our findings demonstrate a high degree of coincidence between the regions associated with a specific Gleason grade and the clusters obtained through hierarchical agglomerative clustering, corresponding to sets of glands with similar morphologies. This segregation was further improved following the CL step. We note an exception to this agreement in a cluster of the feature space where both glands of Gleason pattern 3+4 and 4+3 are found. These intermediate-risk Gleason grades are often difficult to distinguish and can lead to disagreement between pathologists (\cite{Allsbrook2001}). Overall, our results suggest a high degree of agreement between the distinctions that histopathologists make between morphologies of different glands and those identified by the machine, thereby confirming that semantic tiling produces meaningful representations of tissue histology.

\begin{figure}
\begin{center}
\includegraphics[width=\linewidth, page=3]{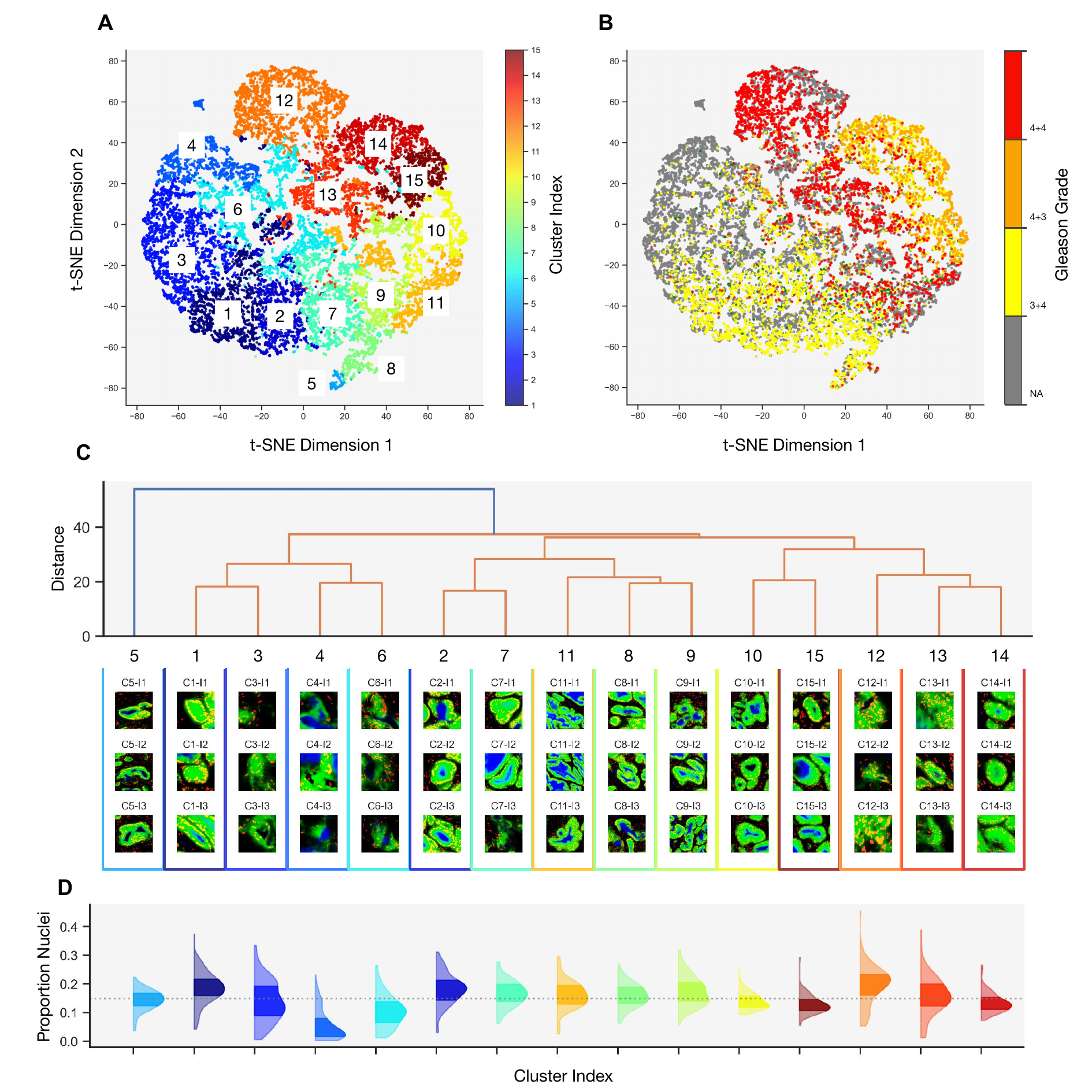}
\end{center}
\caption{
\textbf{Phenotyping of glandular structures and comparison with pathologists’ annotations.} Panel A shows a two-dimensional projection of gland embeddings of glands extracted from FFPE samples from TCGA-PRAD. Each dot corresponds to a gland and the colours indicate the 1–15 morphological clusters identified through HAC. Panel B displays the corresponding Gleason grade associated with these glands. Panel C shows randomly selected examples of tiles from each cluster (black: background, green: epithelium, blue: lumen, and red: nuclei), arranged according to a dendrogram of cluster centroids, which groups the most morphologically similar clusters. Panel D shows a comparison of the proportion of nuclei in each tile across the different clusters. 
}
\end{figure}

\centersection{HIT can improve the performance of CNN-based MIL}

The agreement found between histopathologists’ annotations and morphological clusters, confirmed our expectation of good performance from downstream pipelines that rely on embedding spaces generated through semantic tiling, especially for predicting disease relapse and EMT-related genes, which are both related to gland morphology.

Fig. 4 presents the effect of the tiling methodology, either grid-tiling (GT) or semantic tiling (ST), on the performance of MIL models for predicting BCR (i.e. disease relapse) from WSIs in the ICGC-C cohort, and copy number gain in EMT-related genes and \textit{MYC} in the TCGA-PRAD cohort. Semantic tiling improved the area under the curve (AUC) by nearly 10\% of detecting copy number gain of EMT-related genes and \textit{MYC} when using CNV backbones (Fig. 4A, EMT-CNV \& MYC-CNV, comparing GT and ST). 

This confirmed our initial hypothesis that providing histological structures meaningful to histopathologists to the MIL models could help with model performance. This could be due to a combination of factors. EMT-related genes disrupt the organisation of the epithelial barriers by removing cell-cell junctions (\cite{Mak_2010}). So, by tiling glands, we increase the propensity of epithelial tissue in the images, relative to stromal tissue, and hence potentially increase the signal-to-noise ratio. In addition, tiling around glands of different sizes, enables the MIL pipeline to aggregate features across multiple scales, which has been shown to improve performance (\cite{Shi_2023b}).

\begin{figure}
\begin{center}
\includegraphics[width=\linewidth, page=4]{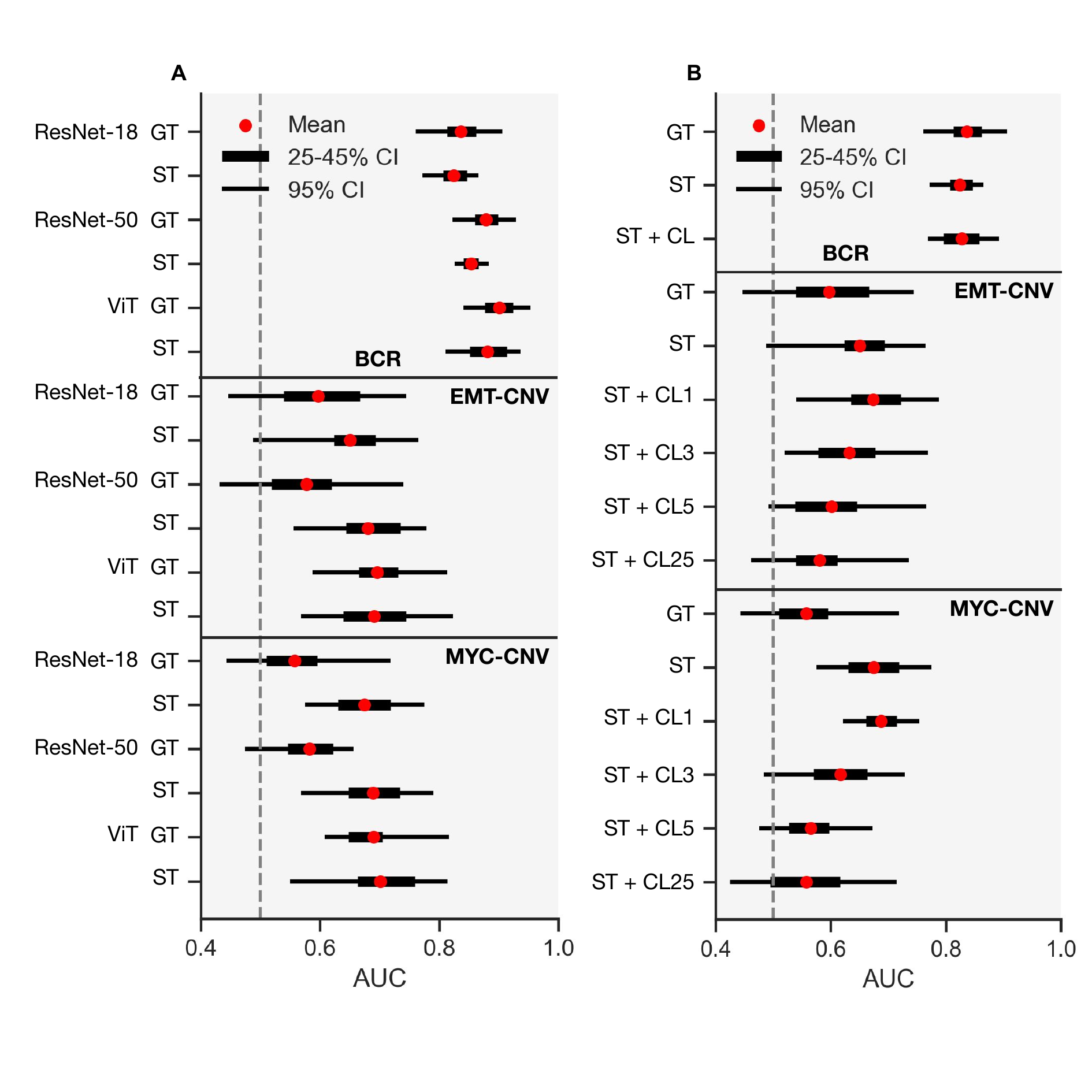}
\end{center}
\caption{
\textbf{Impact of HIT and contrastive learning on performance of multiple-instance learning (MIL) models for relapse and CNV detection.} Panel A shows a benchmark of the accuracy (AUC) of a MIL pipeline in detecting BCR (biochemical recurrence, or relapse) in FFPE patient samples from the ICGC-C cohort, EMT genes copy number variation (EMT-CNV) and copy number gain of the \textit{MYC} gene in OCT patient samples from the TCGA-PRAD cohort. The y-axis labels indicate which method was used for tiling (GT: grid tiling, or ST: semantic tiling a.k.a. HIT) and which backbone was used for compression (ResNet-18, ResNet-50, or ViT). Panel B shows the effects of increasing the number of epochs of contrastive learning on AUC for the ResNet-18 backbone on the three MIL tasks. CL1–25 indicate the number of epochs of contrastive learning fine tuning. In both panels, the aggregator architecture used was CLAM (\cite{Lu_2021}).
}
\end{figure}

\centersection{Glandular structures are sufficient for predicting relapse and mutations}

Surprisingly, ViT backbones performed equally-well on both grid- and semantic tiling, on par with performance of the CNN backbones using HIT. This suggests that ViT models already have an excellent capacity to focus on most meaningful features of the images, due to the self-attention mechanism and capacity for long range associations, which makes them more robust to differences in tiling scheme than CNNs that rely on local feature aggregations (\cite{Dosovitstkiy_2021}). We also found no substantial differences in the performance of MIL models in predicting BCR from grid or semantic tiling (Fig. 4A, BCR). This suggests that the embedding spaces obtained from grid tiling and semantic tiling both contain the necessary information to perform this task. 	

While our favoured hypothesis was that analysing key histological structure would improve downstream tasks performance, the reverse was also possible, namely that deliberately discarding tissue regions that were deemed less relevant might lead to an overall loss of information and therefore to a decrease in performance. The fact that we do not find this strongly suggests that focussing on glandular structures is sufficient for predicting disease relapse and genetic alterations in prostate cancer, thereby confirming the validity of Gleason scoring which relies on the assessment of gland morphology. Furthermore, where performance does not improve following HIT, we argue that a model that requires less information than another should be preferred on the basis of parsimony, all else being equal. We hence view HIT as a more streamlined approach to digital histopathology. 

\centersection{Gland architectures may be more informative as a continuum}

We found limited evidence that using CL to fine-tune the encoders could further improve the performance of MIL pipelines for predicting BCR and CNVs. Marginal improvements can be achieved after one epoch of training, but further training rapidly deteriorates the performance (Fig. 4B). This suggests that these approaches can lead to a loss of diversity in the feature space, which contrasts with previous studies that showed a clear benefit of CL (e.g. \cite{Wang_2022, Liu_2024}). Further work on integrating these approaches with CL should clarify this issue.
	
CL had a clear impact on the embedding space by aggregating patches and delineating groups of similar glands. The fact that this leads to a deterioration of performance suggests that a key factor underpinning model predictions is the possibility to draw inference from a continuous distribution of gland morphologies. This may also be one of the limitations of the Gleason scoring system, which stems from the arbitrary discretisation of this continuum. We propose that while CL is useful for visualising and interpreting clusters, it may be a counterproductive step, in terms of performance, in cases where the information is distributed over a continuum.

\centersection{Gland-centric phenotyping can improve explainability in digital pathology}

Beyond the benefits in performance, tiling according to histological structures can improve the interpretability of cancer phenotyping in digital histopathology. We show in Fig. 5 how the combination of the clustering and MIL pipelines enable the phenotyping of gland morphologies associated with EMT. EMT evidence, indicated by EMT-positive instances, was concentrated in three main areas of the embedding space (Fig. 5B). We found that clusters 2, 13, 9, and 11 contained the greatest number of EMT-positive instances (Fig. 5D). Apart from cluster 2, these clusters displayed a relatively higher density of cells (Fig. 5D). We found similar results for the phenotyping of glands associated with BCR (Fig. S2). Overall, evidence for BCR was more localised in the embedding space (Fig. S2B). We found that cluster 15 and 6 contained the most BCR-positive instances, with cluster 15 containing a higher density of cells, compared to cluster 6 which was the cluster with the lowest cell density (Fig. S2D). Overall, the glands from these clusters were smaller and contained a higher number of nuclei in line with the idea that more aggressive cancer forms can be associated with local cell proliferation through de-differentiation in intermediate cell stages (\cite{Byles_2012}). This further aligns well with the gland phenotypes identified in clusters mapping to higher Gleason grades (Fig. 3D, cluster 13 \& 12).

Another aspect of our approach that improves interpretability of model outputs is that it provides attention maps at the gland-level instead of arbitrarily tiled patches, hence flagging specific structures rather than general regions of interest. Therefore, both histopathologists and MIL pipelines operate on the same histological structures, making it easier to match with clinically relevant tissue properties. We show in Fig. 6 a few examples of the application of the pipeline to WSIs to demonstrate how it can be used to identify clusters of glands, and detect evidence of BCR and EMT.  

\begin{figure}
\begin{center}
\includegraphics[width=\linewidth, page=5]{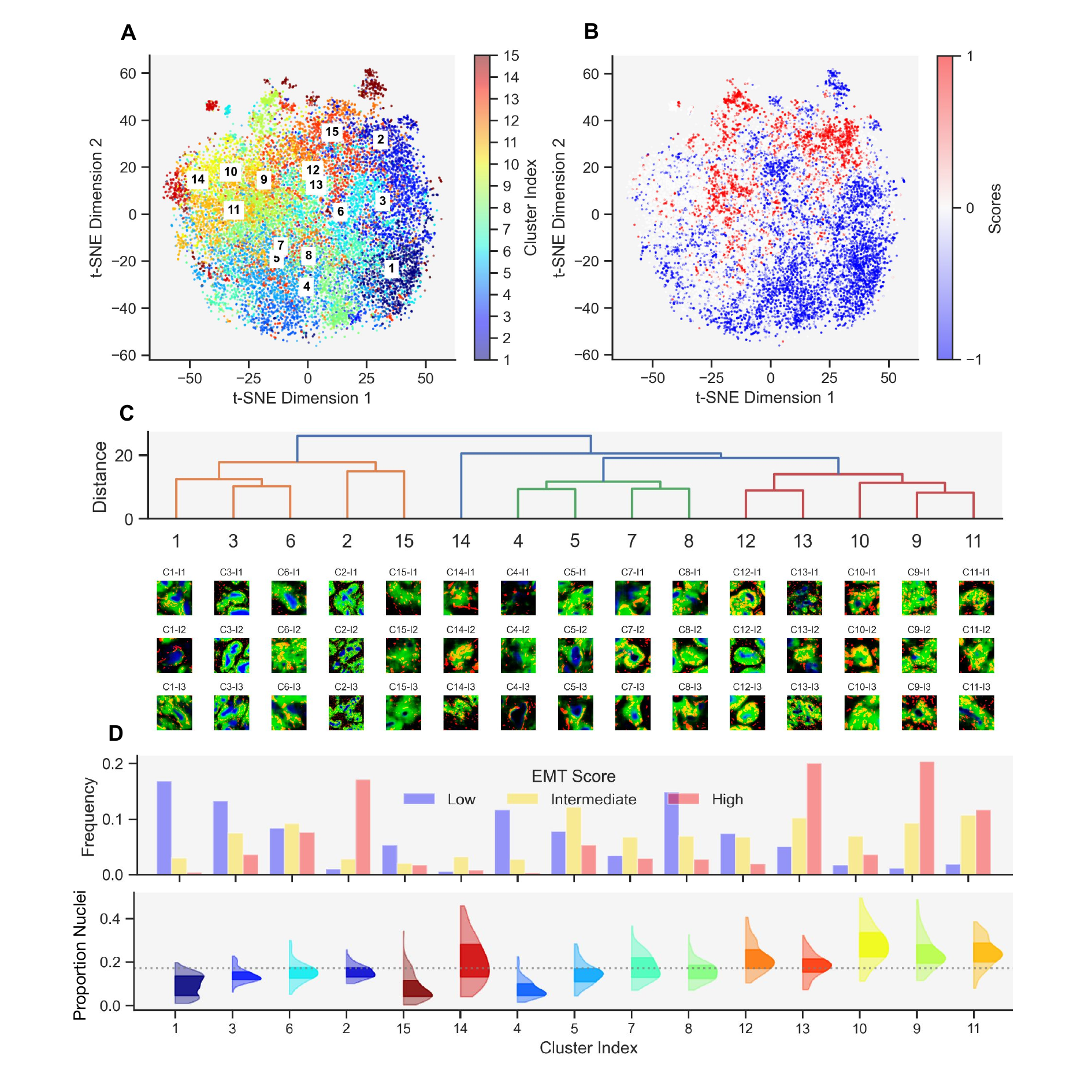}
\end{center}
\caption{
\textbf{Phenotyping of glands associated with copy number gain of EMT-related genes.} Panel A displays a two-dimensional projection of the embeddings of glands extracted from OCT samples of the TCGA-PRAD cohort. The colour indicates the index of the 1–15 clusters established through HAC. Panel B shows the corresponding prediction-attention-weighted scores for each instance (i.e. EMT scores), where blue and red dots are instances that decrease or increase the probability of EMT gain, respectively. White dots are intermediate instances that neither increase nor decrease the probability. Panel C shows examples from each identified cluster (with epithelium in green, lumen in blue, and nuclei in red). Relationships between morphologically similar clusters are indicated by a dendrogram. Panel D shows the frequency of instances with low (blue), intermediate (yellow), and high (red) EMT scores, and the corresponding proportion of nuclei, within each cluster.
}
\end{figure}

\begin{figure}
\begin{center}
\includegraphics[width=\linewidth, page=6]{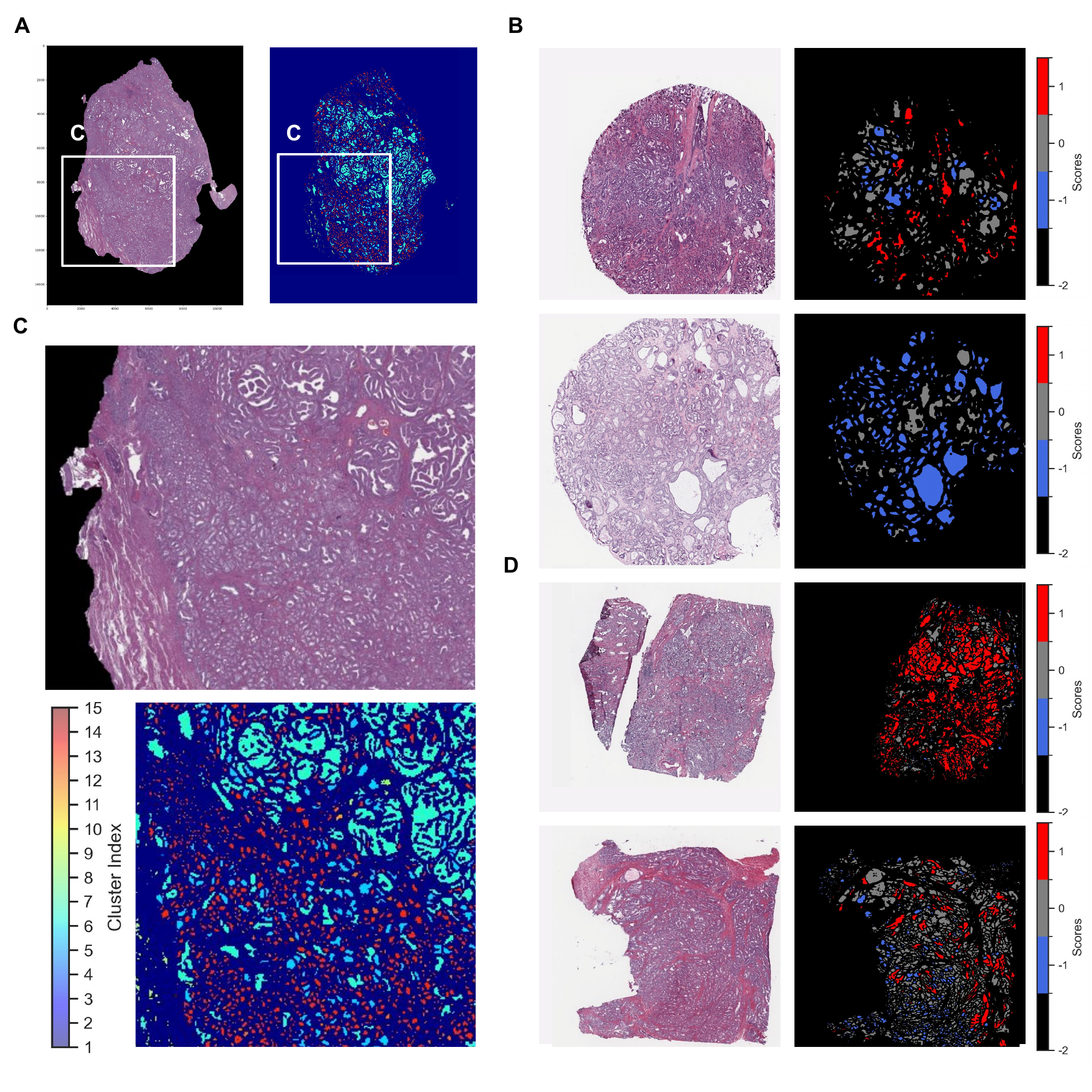}
\end{center}
\caption{
\textbf{Example of WSIs and gland-level predicted clusters and BCR and EMT scores.} Panel A shows an example of WSI and resulting slide-level cluster predictions. Panel C shows the detail of the cluster prediction map, where each gland is associated with a cluster index (1–15). Panel B displays the input WSIs along with gland-level BCR (i.e. relapse) scores (blue: low risk, red: high risk) for a BCR positive patient (top) and negative (bottom). Panel D displays the input WSIs of an EMT positive (top) and negative (bottom) case with gland-level predictions of risk of EMT gain (blue: low risk, red: high risk).
}
\end{figure}

\section{Discussion}

We propose HIT, a new approach to MIL and cluster-based phenotyping which divides WSIs into tiles according to histological structures detected through semantic segmentation. This methodology reduces the overall amount of information passed to downstream pipelines, while increasing its quality and interpretability by preserving the histology of the slides. We showed that this could improve the performance of detection of CNV in oncogenic and EMT-related genes in prostate cancer by CNN-based MIL. Interestingly, the fact that performance was unchanged following HIT for ViT architectures and for predicting BCR showed that glandular structures contain sufficient information to perform these tasks.  The dual benefit of HIT is that it maintains a direct link between patches and histological structures, which facilitates the phenotyping of cancer by explicitly flagging specific gland architectures that associate with more aggressive disease, while enabling more efficient computations at the feature extraction stage due to a lower number of gland instances compared to grid tiling. Overall, this approach can change the way we analyse WSIs in digital pathology, by targeting explicitly biologically meaningful histological structures, we can improve computational efficiency and interpretability of our algorithms without trading-off accuracy.

While we focussed on prostate cancer, our approach is generalisable to other tissue types. Glandular structures, in particular, are common in cancers such as breast, colorectal, and pancreatic, making our approach broadly applicable across multiple types of cancer. Furthermore, our HIT stage targets agglomerations of epithelial cells, which is a general histological feature present in all tissue containing epithelial barriers. This means that this approach can be used even on tissues which have no clear glandular structures, as long as they contain epithelial cells. Other histological structures can also be targeted, for instance blood vessels or nerves, which have been found to be important for diagnosis and prognosis (\cite{Magnon_2013}). Overall, this approach is highly flexible and generalisable to target different cancers by targeting ubiquitous histological structures.	

While we did not find evidence of detrimental effects of discarding tissue areas that did not contain glands in this study, there are cases where this may lead to a loss of crucial information. Several studies have pointed to a relationship between reactive stroma and prognosis in prostate cancer (\cite{Ayala_2003, Yanagisawa_2008}). In addition, in EMT, most advanced stages of progression lead to de-differentiation of epithelial cells either in an intermediate stage, or in a mesenchymal stage  (\cite{Byles_2012}). Finally, higher Gleason patterns are associated with glands that have collapsed into small sub-glands with a much less clear epithelial boundary (as seen in \cite{Nguyen_2014}). These histological structures formed by cells with a less-well defined epithelial phenotype may be difficult to detect with our epithelial segmentation approach. More generally, cancer forms that lead to less well-defined histological structures may not be well captured by HIT. More investigations are hence required to clarify the relative importance of well-defined histological structures compared to less conspicuous features.	

\section{Methods}

\noindent
\textbf{Overview.}
The proposed methodology first took as input WSIs at 40x and produced slide-wide prediction masks for the epithelium and lumen of prostate glands, which were then used to extract images of individual glands. In a second step, our methodology leveraged these gland-level patches and CL to learn fine morphological features of prostate glands and identify clusters of glands with similar morphologies. Finally, raw or refined features were used as inputs to a MIL pipeline to perform binary classification tasks: prognosis and detection of CNVs.

\vspace{0.2cm}
\noindent
\textbf{Cohorts.}
In this study, we used WSIs from three cohorts of patients: ProMPT (Prostate Mechanisms of Progression and Treatment), ICGC-C (International Cancer Genome Consortium), and TCGA-PRAD (The Cancer Genome Atlas). 

First, we used the ProMPT cohort to train the segmentation model to extract individual prostate glands.  
We included a subset of 24 H\&E-stained (Haematoxylin and Eosin) FFPE (Formalin-Fixed Paraffin-Embedded) WSIs, corresponding to 24 different patients from a digitised version of the cohort (\cite{Haghighat_2022}). The tissue samples corresponded to pucks obtained from radical prostatectomy (RP) specimens, which were then prepared and sliced for imaging through light microscopy. The slices were digitised in the ‘.isyntax’ format using a Philips Pathology Scanner. Once the slices were digitised, expert histopathologists identified 137 regions of interest (ROIs) where they annotated at the pixel-level the epithelial boundary and lumen of prostate glands. The ROIs were selected to span a range of gland morphologies representative of each Gleason pattern. 

Second, we used the locally-processed ICGC-C cohort for assessing the influence of HIT in predicting prostate cancer relapse. This dataset comprised 516 H\&E-stained WSIs, obtained from 95 patients. 381/516 samples were tissue diagnosed with malignant tumour, and 98/516 samples came from patients diagnosed with post-treatment prostate cancer relapse.

Finally, we used the publicly accessible TCGA-PRAD cohort to assess the benefit of HIT in detecting CNV in the oncogene \textit{MYC} and EMT-related genes. TCGA-PRAD is an open dataset which combines FFPE WSIs from RP specimens and bulk transcriptomics and genomics (\cite{Tomczak_2015}). We focussed on open cases with OCT WSIs from primary tumour sites with available bulk genomics and CNV profiles. Data were obtained from the GDC portal by selecting 234 cases with a gene-level CNV profile obtained following the ABSOLUT LiftOver workflow and linked to OCT WSIs from primary tumour sites. In addition, we randomly selected and downloaded 10 high quality FFPE WSIs from the TCGA-PRAD dataset for Gleason grade annotation by histopathologists and comparison with gland clusters identified using HIT.

\vspace{0.2cm}
\noindent
\textbf{Target labels.}
We applied HIT to two key binary classification tasks based on WSI analysis: (1) quantifying the risk of cancer relapse to inform disease prognosis, and (2) detecting genetic alterations indicative of aggressive cancer forms. For the first task, disease relapse was determined by assessing biochemical-recurrence (BCR) at 5-years clinical follow-up. We used BCR for lack of information regarding the development of distant metastasis and prostate-specific mortality in our datasets. For the second task, we focussed on \textit{MYC}, which is a well-known oncogene (Weinberg 2014), and EMT-related genes \textit{ZEB1}, \textit{ZEB2}, \textit{SNAI1}, \textit{SNAI2}, and \textit{CDH1}, which are known to destabilise epithelial barriers (\cite{Mak_2010}), the initial step in the formation of metastases. Due to the sparsity of the CNV data, and the functional redundancy between EMT-related genes, we combined the EMT-related CNVs into a binary EMT score set to 1 if at least one copy number gain in any of the EMT-related genes was detected in a patient sample, and 0 otherwise.

\vspace{0.2cm}
\noindent
\textbf{Segmentation dataset preparation and augmentation.}
The starting point of HIT was the segmentation of clinically-relevant tissue structures, such as glands in the case of prostate cancer. We thus created a dataset suitable for training gland segmentation models. Histopathologists annotated glands at the pixel level across the 137 ROIs, creating a dataset for training the segmentation model (Fig. 2B). 10 ROIs were selected at random reserved for post-training tests, while 118 and 9 of the remaining ROIs were randomly assigned to a training and validation set, respectively. Random patches (1024 $\times$ 1024 $\times$ 3) were extracted from these ROIs and augmented tenfold through rotations, flips, and colour alterations to enhance the diversity and robustness of the training dataset.

\vspace{0.2cm}
\noindent
\textbf{Segmentation model architecture.}
The segmentation model employed was a U-Net convolutional encoder-decoder combined with a transformer module operating on the innermost feature map, allowing for context-aware segmentation. The model accepted input patches (1024 $\times$ 1024 $\times$ 3) at the highest resolution (40X) and predicted the probability of three pixel-level classes, stroma or background, epithelium, and lumen, on a per-patch basis (Fig. 2B). This output was then divided into compartment-specific prediction masks by applying a threshold set to 0.5 by default. This resulted in binary masks indicating the presence or absence of a given tissue type at the pixel level. Details of the number of convolutional blocks and number of filters within each block are provided in the code. We also implemented a standard U-Net to assess the benefit of including the transformer module in the architecture. We did not find substantial differences in the performance of the U-Net with or without the transformer. To obtain the gland masks, we joined the predictions of the two models by multiplying the predicted class probabilities. 

\vspace{0.2cm}
\noindent
\textbf{Segmentation models training.} 
The models were trained on the training set and evaluated on the validation set for each epoch. We used the cross-entropy loss, which evaluates the pixel-wise alignment between the predicted probability distribution and the ground truth class distribution for each class. The models were trained for a maximum of 90 epochs, as no substantial improvement in the validation loss was observed beyond this point. We exhaustively tested different combinations of batch sizes (1 to 4), learning rates (1e-3 or 1e-4), and optimisers (Adam or SGD) to identify the best validation loss. The best hyperparameters identified for this dataset were a batch size of 4, learn rate of 1e-4, and SGD as the optimiser. After training, the performance of the model on unseen data was obtained by predicting the gland masks in the test set. The quality of the segmentation masks for each compartment of the tissue was computed as the Dice score, which corresponds to the agreement between the predicted mask and the ground truth (i.e. twice the intersection divided by the union).

\vspace{0.2cm}
\noindent
\textbf{Extraction of glands from WSIs.}
Once the segmentation framework was trained and tested, we applied it to extract glands from the WSIs of the ICGC-C and TCGA-PRAD cohorts. Because glands can extend beyond the input patches of the U-Nets, we extracted partially overlapping patches at the highest resolution (40X) following a grid pattern to ensure comprehensive coverage. For each patch, the model generated patch-level segmentation masks for each tissue compartment. Patch-level predictions were then downsampled by 2 or 4 (depending on memory constraints) before being collected and reassembled to obtain slide-level prediction masks. The slide-level prediction masks of the epithelium and lumen compartments were combined to create a gland prediction mask. Using contour finding methods and object extraction methods (the function \textit{connectedComponents} in OpenCV) we extracted each individual gland identified by the model. Finally, gland-level images were exported and saved locally for further analysis (Fig. 3A–F).

\vspace{0.2cm}
\noindent
\textbf{Generation of gland embeddings.}
Traditionally, the use of AI in digital pathology, either for predicting diagnosis (i.e. the nature and current state of a pathology) or prognosis (i.e. likely course of evolution of the disease) from WSIs, relies on a pre-processing step that consists of extracting tissue patches which are then compressed into embedding vectors using an image encoder pre-trained on another dataset. This compression step, also referred to as feature extraction, enables faster computations and facilitates downstream analysis. We hence prepared each extracted gland image for feature extraction. The patches were re-sized to 224 $\times$ 224 $\times$ 3 and normalised. We used three different encoder architectures: two CNN-based architectures (ResNet-18, ResNet-50), and a vision transformer (ViT, \cite{Dosovitstkiy_2021}). To generate embeddings, we selected a given architecture, processed all patches for a given cohort, and collected the output embedding vectors. The feature vector of each patch had a dimension of 512. This resulted in the generation of two embedding datasets, one for the ICGC-C cohort (516 samples) and another for the TCGA-PRAD cohort (234 samples).

\vspace{0.2cm}
\noindent
\textbf{Fine-tuning encoders with self-supervised CL.}
We also assessed whether fine-tuning the encoders on gland morphologies with CL could further enhance performance. Given that this was not the primary focus of the study, we only considered fine-tuning ResNet-18 for simplicity. Individual gland images were hence pre-processed according to the requirements of the ResNet-18 architecture (e.g. normalisation followed by standardisation of images). All images were resized to 224 $\times$ 224 $\times$ 3. This ensured that relative gland size could only be inferred from morphology and not from pixel area, thereby reducing the likelihood of obtaining purely size-based gland features.
We then trained the ResNet-18 encoder to recognise subtle differences in the morphological features of glands through CL. We used triplet loss (\cite{Hermans_2017}), which encourages the model to minimise the distance between an image and a transformed representation of the same image while maximising the distance with other images. We also tested SimCLR loss but found no substantial difference in preliminary results, so we focussed on triplet loss for the remainder of the study.  Overall, CL fine-tuned the encoder to group similar gland structures and separate dissimilar ones in a reduced feature space (Fig. S1). We used a margin parameter of 0.75, a projection dimension of 128, a batch size of 128, and a 70–20–10\% train-validation-test split. Image transformations included random horizontal and vertical flips, as well as colour jitter. We tested different levels of feature refinement by saving checkpoints after 1, 3, 5, and 25 epochs (Fig. 4B). 

\vspace{0.2cm}
\noindent
\textbf{Unsupervised clustering and classification of gland morphologies.}
Once the embeddings were generated by the different encoders, we used HAC to identify gland clusters in the embedding space. HAC proceeds by iteratively grouping the two closest points in the latent space until a single root cluster is obtained. This forms a dendrogram of all data points, which can be cut at different levels to obtain different numbers of clusters.  We performed HAC on the embeddings with an increasing number of clusters (i.e. 1–20) and computed the within-cluster and between-cluster variance of the embedding vectors. We selected the number of clusters that maximised the ratio of between-cluster to within-cluster variance ratio (Fig. S2B–C). Following this approach, we established the cluster identities of all glands in our datasets. We also computed a dendrogram of cluster centroids to understand which clusters were most similar, and whether there were meta-clusters in our datasets, i.e. groups of clusters (Fig. 3C, 5C, S2C).
To classify new instances, e.g. glands from newly digitised slides, we then trained a k-nearest neighbours classifier to predict the cluster identity of glands based on their distance to neighbour glands with known cluster identity. We chose k = 3 for the sake of simplicity, as this is purely for visualisation of clusters in new slides. None of the analyses presented in this manuscript depended on this classification step.

\vspace{0.2cm}
\noindent
\textbf{Comparison of clusters to histopathologists’ annotations.}
We compared the clusters identified by HAC to gland morphologies corresponding to the different Gleason patterns identified by histopathologists. To do this, we selected 4/10 WSIs in the FFPE TCGA-PRAD cohort which presented a wide range of gland morphologies and had histopathologists divide the samples into regions of different Gleason grades (i.e. Gleason 3+3 and lower, 3+4, 4+3, 4+4 and higher). We applied our segmentation and clustering methodology to characterise the clusters and gland morphologies found within the different Gleason regions. 

\vspace{0.2cm}
\noindent
\textbf{Phenotyping of gland clusters.}
To further understand the morphologies of glands contained in the different clusters, we computed hand-crafted features, i.e. quantitative measures relating to the geometric properties of the glands. We computed the area of each gland (in pixels) as a proxy for gland size. We used the different segmented compartments for each gland to compute the relative luminal, epithelial, and stromal areas. In addition to \textit{GlandSeg}, we also trained a nuclei segmentation model, hereafter referred to as \textit{NucleiSeg}, on MoNuSeg, a publicly accessible dataset for training nuclei segmentation models (\cite{Kumar_2017}). We applied \textit{NucleiSeg} to the extracted gland images. This allowed us to determine relative nuclear area in the images (in pixels) as a proxy for cellular density within the glands. By combining nuclei segmentation masks with tissue compartment masks from \textit{GlandSeg} (i.e. stroma, epithelium, and lumen) we were able to compute epithelial and stromal cell densities. In this manuscript, we focussed on epithelial nuclei density as the main feature of interest. This is because the local proliferation of cancer cells is characterised by disorganised and dense cell aggregates within glands (\cite{Nguyen_2014}). 

\vspace{0.2cm}
\noindent
\textbf{Preparation of embeddings for MIL.}
In addition to cluster-based phenotyping, we assessed the usefulness of HIT for MIL classification of BCR and CNVs. To format embeddings for MIL, gland embeddings originating from the same WSI were grouped into the same bag. This resulted in a table with dimensions N slides $\times$ N instances $\times$ N features. To limit the impact of the variation of the number of instances between slides, we set the maximum number of instances per bag to 1,000. Instances were randomly selected until this maximum was reached. Slides with fewer instances were padded with null embedding vectors. We verified that our results were robust to either increasing or relaxing this constraint by re-running the analyses with the maximum number of instances set to 100 and 2,000.

\vspace{0.2cm}
\noindent
\textbf{MIL models.}
MIL relies on aggregator architectures (i.e. shallow classification heads) to collect information across all instances contained in a bag to make a prediction at the bag level (\cite{Isle_2018}). Attention-based MIL is the baseline architecture for most pipelines in digital pathology (\cite{Isle_2018}). It uses attention pooling to increase the importance of tissue areas containing relevant information while reducing that of others. In this study, we used two attention-based MIL architectures. The first was CLAM (\cite{Lu_2021}), a widely used and validated architecture. The other architecture, named PAW-MIL (prediction-attention-weighted MIL), was derived from an implementation that introduces prediction-attention weighting at the pooling stage (\cite{Bonnaffe_2023}). It uses two parallel multi-layer perceptrons that separate the attention scores, namely the importance, of instances from their effects or contribution, either negative or positive, to the bag-level prediction. A similar implementation, Attri-MIL, demonstrated the benefit of this decomposition for performance (\cite{Cai_2025}).

\vspace{0.2cm}
\noindent
\textbf{Training aggregators and hyperparameter selection.}
To train the MIL models, we relied on a grouped stratified k-fold cross-validation scheme, whereby bags of instances belonging to the same case ID (i.e. the grouping variable) were divided into training, validation, and test sets. Splits were performed so that the proportion of negative and positive labels was preserved across folds and all samples belonging to the same case ID were contained within the same set. The test set was reserved first, following an 80–20\% split of the entire dataset. Then, 5 train–validation folds were generated following another 80–20\% split of the remaining data. The model was then trained on each training set and performance assessed on the corresponding validation set. Hyperparameters were sequentially altered: the batch size was increased from 4 to 8, the learn rate from 1e-4 to 1e-3, and the number of epochs from 8 to 16 to 32. Training was repeated 3 times to assess training consistency for each combination of hyperparameters. The hyperparameter set with the highest mean validation AUC across the five train–validation folds was selected as optimal. The entire process was then repeated for the five possible test splits to assess robustness to sampling. Model performance was assessed by computing the AUC across the 5 test folds and 3 repeated training runs.

\vspace{0.2cm}
\noindent
\textbf{Computing gland-level prediction scores and slide-level prediction-attention maps.}
The motivation behind using PAW-MIL in this study was that it enabled the computation of the effect of each instance on the slide-level prediction, which allowed us to categorise glands as positive or negative evidence. Our MIL training scheme, which used 5 test folds and 3 repeat optimisation runs per fold, generated an ensemble of 15 models from which predictions were derived. This allowed us to obtain an ensemble of predictions for each instance and determine whether the PAW score for that instance was significantly greater or lower than 0, defined as having all prediction scores from the model ensemble above or below 0, respectively. For example, a gland patch received an EMT score of 1 if all ensemble predictions were above 0, and vice versa. Otherwise, the score was deemed intermediate, namely that it was neither positive or negative.
Prediction-attention maps were computed by collecting PAW scores across all slide instances and visualising them in the original tissue location. The difference with standard grid-tiling is that the attention scores are at the level of the entire histological structure. Therefore, the scores applied to all pixels belonging to the extracted glands. The prediction-attention maps thus highlighted the most important glandular structures for the prognosis and CNV detection task. 

\vspace{0.2cm}
\noindent
\textbf{Benchmark study.}
To determine the effect of HIT on the performance of MIL classifiers for BCR and CNV detection, we designed a benchmarking experiment. We compared the performance of a MIL model trained on tile sets obtained either through grid- or histology-informed tiling and encoded by ResNet-18, ResNet-50, or ViT. This produced 6 different embedding sets for each cohort. This allowed us to test the robustness of the effect of the tiling method to the choice of encoder architecture. In addition, we assessed potential improvements of performance after additional epochs of fine-tuning with CL. For CL, we only tested the most simple architecture, the ResNet-18, and did not assess more complex architectures, as no clear benefits were observed in this initial stage. 

\vspace{0.2cm}
\noindent
\textbf{Hardware and code.}
The methodology and models were implemented in Python using the deep learning library PyTorch. Training of the segmentation models, the CL step, and MIL models were performed locally on a laptop (MacBook Pro M1 Max, 16-inch, 2021).


\vspace{0.2cm}
\noindent
\textbf{Data availability.}
The datasets from the ProMPT cohort containing the H\&E-stained tissue regions of interest and corresponding gland annotations that we used for the training of the segmentation models will be available in a dedicated repository on Zenodo (\url{https://zenodo.org/}). 
The dataset from the ICGC-C cohort that we used to predict BCR from WSIs can be provided upon reasonable request. 
The dataset from the TCGA-PRAD cohort is publicly available at \url{https://portal.gdc.cancer.gov/}.
MoNuSeg, the dataset that we used to train the nuclei segmentation model is available in the publication repository of the original paper (\cite{Kumar_2017}).

\vspace{0.2cm}
\noindent
\textbf{Code availability.}
All code will be made available in a GitHub repository (\url{https://github.com/willembonnaffe/CancerPhenotyper}) and archived on Zenodo along with the datasets.

\centersection{Acknowledgments}

Clare Verrill is supported by the NIHR Oxford Biomedical Research Center (BRC4-SITE; sub-theme 5—pre-operative optimisation and enhanced recovery). The views expressed are those of the author(s) and not necessarily those of the NHS, the NIHR, or the Department of Health. PathLAKE Centre of Excellence for Digital Pathology and Artificial Intelligence, funded by the Data to Early Diagnosis and Precision Medicine strand of the HM Government's Industrial Strategy Challenge Fund, managed, and delivered by Innovate UK on behalf of UK Research and Innovation (UKRI). Grant Ref: File Ref 104689/Application Number 18181. An extension and expansion of the program is also funded as ‘PathLAKE plus’. Views expressed are those of the authors and not necessarily those of the PathLAKE Consortium members, the NHS, Innovate UK, or UKRI. The national ProMPT (Prostate cancer Mechanisms of Progression and Treatment) collaborative (Grant G0500966/75466) supported sample collections. The University of Oxford sponsors ProMPT. Additional to the named authors, the CRUK ICGC Prostate Group also contains the following members: Adam Lambert, University of Oxford, Oxford, UK; Anne Babbage, Hutchison/MRC Research Centre, Cambridge University, Cambridge, UK; Claudia Buhigas, Norwich Medical School, University of East Anglia, Norwich, UK; Dan Berney, Department of Molecular Oncology, Barts Cancer Centre, London, UK; Nening Dennis, Royal Marsden NHS Foundation Trust, London and Sutton, UK; Sue Merson, The Institute Of Cancer Research, London, UK; Alastair D. Lamb, Nuffield Department of Surgical Sciences, University of Oxford, Oxford, UK; Adam Butler, Cancer Genome Project, Wellcome Trust Sanger Institute, Hinxton, UK; Anne Y. Warren, Department of Histopathology, Cambridge University Hospitals NHS Foundation Trust, Cambridge, UK; Vincent Gnanapragasam, Department of Surgical Oncology, University of Cambridge, Addenbrooke’s Hospital, Cambridge, UK; G. Steven Bova, Prostate Cancer Research Center, Faculty of Medicine and Health Technology, Tampere University, Finland; Christopher S. Foster, HCA Laboratories, London, UK; David E. Neal, Department of Surgical Oncology, University of Cambridge, Addenbrooke’s Hospital, Cambridge, UK; Yong-Jie Lu, Centre for Molecular Oncology, Barts Cancer Institute, Queen Mary University of London, London, UK; Zsofia Kote-Jarai, The Institute of Cancer Research, London, UK; Robert G. Bristow, Manchester Cancer Research Centre, Manchester, UK; Andy G. Lynch, School of Medicine/School of Mathematics and Statistics, University of St Andrews, St Andrews, UK; Daniel S. Brewer, Norwich Medical School, University of East Anglia, Norwich, UK; David C. Wedge, Manchester Cancer Research Centre, Manchester, UK; Rosalind A. Eeles, The Institute of Cancer Research, London, UK; Colin S. Cooper, Norwich Medical School, University of East Anglia, Norwich, UK. This work was supported by grant MA-ETNA19-005 'Major Awards, Existing Trials: New Answers' funded by Prostate Cancer UK.

\centersection{Author contributions (CRediT)}

\textbf{Willem Bonnaffé:} Conceptualisation, Methodology, Software, Validation, Formal analysis, Investigation, Resources, Data Curation, Writing - Original Draft, Writing - Review \& Editing, Visualisation, Project Administration; 
\textbf{Yang Hu:} Conceptualisation, Methodology, Writing - Review \& Editing; 
\textbf{Andrea Chatrian:} Conceptualisation, Methodology, Resources, Data Curation; 
\textbf{Mengran Fan:} Data Curation; 
\textbf{Stefano Malacrino:} Software, Resources, Data Curation; 
\textbf{Sandy Figiel:} Writing - Review \& Editing; 
\textbf{CRUK ICGC Prostate Group:} Resources, Data Curation; 
\textbf{Sirinavasa R. Rao:} Conceptualisation, Methodology, Writing - Review \& Editing; 
\textbf{Richard Colling:} Resources;
\textbf{Richard J. Bryant:} Writing - Review \& Editing; 
\textbf{Freddie C. Hamdy:} Project Administration, Funding acquisition; 
\textbf{Dan J. Woodcock:} Conceptualisation, Methodology, Investigation, Resources, Data Curation, Writing - Review \& Editing, Supervision, Project Administration, Funding acquisition; 
\textbf{Ian G. Mills:} Conceptualisation, Methodology, Investigation, Resources, Writing - Review \& Editing, Supervision, Project Administration, Funding acquisition; 
\textbf{Clare Verrill:} Conceptualisation, Methodology, Investigation, Resources, Data Curation, Writing - Review \& Editing, Supervision, Project Administration, Funding acquisition; 
\textbf{Jens Rittscher:} Conceptualisation, Methodology, Investigation, Resources, Data Curation, Writing - Review \& Editing, Supervision, Project Administration, Funding acquisition.

\printbibliography

\appendix
\newpage
\setcounter{figure}{0}
\renewcommand{\thefigure}{S\arabic{figure}}
\setcounter{table}{0}
\renewcommand{\thetable}{S\arabic{table}}

\begin{figure}
\begin{center}
\includegraphics[width=\linewidth, page=7]{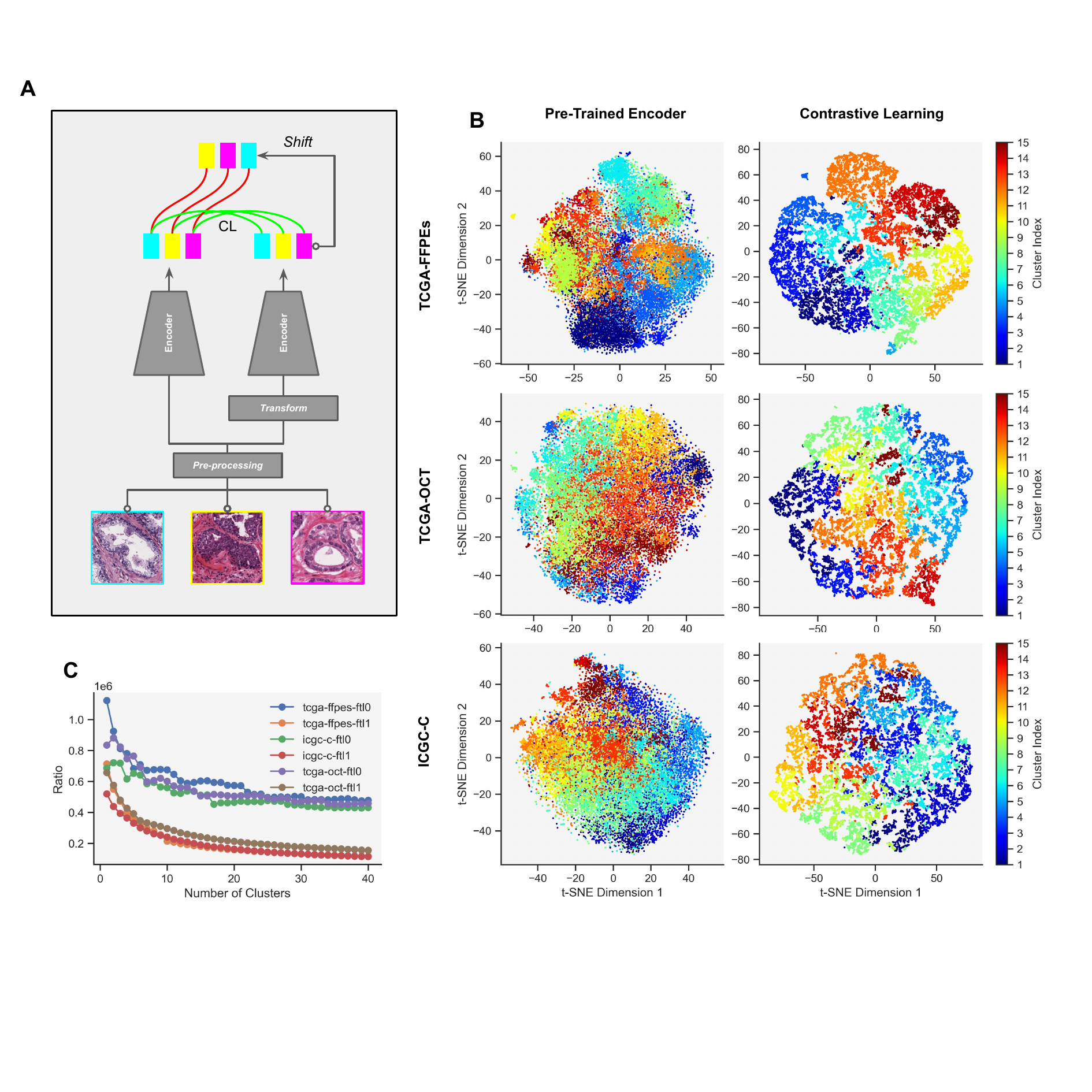}
\end{center}
\caption{
\textbf{Conceptual overview of contrastive learning approach and effect on embedding space structuration.} Panel A shows the contrastive learning algorithm using triplet loss to fine-tune the encoder. Panel B displays the embedding space generated by either the naive ResNet-18 encoder (Pre-Trained encoder) or the fine-tuned encoder (Contrastive Learning) for the three different datasets (FFPEs- and OCT-TCGA, and ICGC-C). Each point corresponds to an embedded gland. Different colours indicate the cluster index obtained through HAC. Panel C shows the change in the intra-to-inter-cluster variance ratio for the three datasets, before and after contrastive learning, noted ftl0 and ftl1 (i.e. fast triplet loss disabled (0) or enabled (1)), respectively. 
}
\end{figure}

\begin{figure}
\begin{center}
\includegraphics[width=\linewidth, page=8]{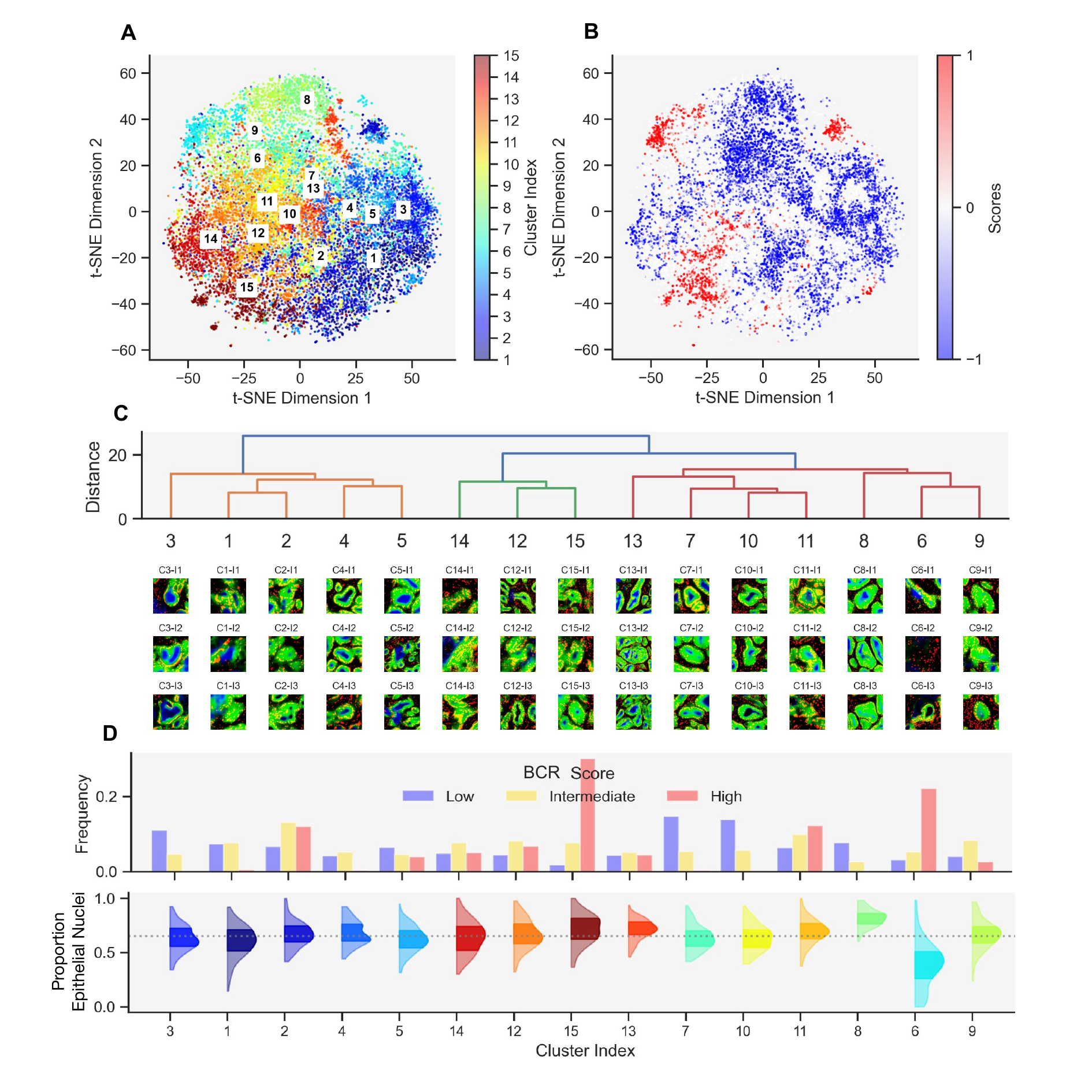}
\end{center}
\caption{
\textbf{Phenotyping of glands associated with BCR.} Panel A displays a two-dimensional projection of the embeddings of glands extracted from WSIs of the ICGC-C cohort. The colour indicates the index of the 1–15 clusters established through HAC. Panel B shows the corresponding prediction-attention-weighted scores for each instance, where blue and red dots are instances that decrease or increase the probability of BCR, respectively. White dots are intermediate instances that neither increase nor decrease the probability. Panel C shows examples from each identified cluster (with epithelium in green, lumen in blue, and nuclei in red). Relationships between morphologically similar clusters are indicated by a dendrogram. Panel D shows the frequency of instances with low (blue), intermediate (yellow), and high (red) BCR scores, and the corresponding proportion of nuclei, within each cluster.
}
\end{figure}

\end{document}